\documentclass{article}

\usepackage[dvipsnames, table]{xcolor}
\usepackage{microtype}
\usepackage{graphicx}
\usepackage{subfigure}
\usepackage{hyperref}

\usepackage[accepted]{icml2025}

\usepackage{amsmath,amsfonts,amsthm,amssymb}
\usepackage{mathtools}
\usepackage{advdate}
\usepackage[mathscr]{euscript}
\usepackage{systeme}
\usepackage{pdfpages}
\usepackage{hyperref}
\usepackage{paralist}
\usepackage{tikz}
\usepackage{colortbl}
\usepackage{xspace}
\usepackage{stfloats}
\usepackage{booktabs}
\usepackage{multirow}
\usepackage{graphicx}
\usepackage{pifont}
\usepackage{wrapfig}    
\usepackage{pgfplots}
\usepackage{siunitx}       % For better number formatting
\usepackage{enumitem}
\usepackage{authblk}
\usepackage{footmisc}

\usepackage[capitalize,noabbrev]{cleveref}

\theoremstyle{plain}

\theoremstyle{definition}

\theoremstyle{remark}

\usepackage[textsize=tiny]{todonotes}
\newcommand{\prompt}{\ensuremath{\mathbf{x}}\xspace}
\newcommand{\xtoken}[1]{\ensuremath{x_{#1}}\xspace}

\newcommand{\response}{\ensuremath{\mathbf{y}}\xspace}
\newcommand{\step}[1]{\ensuremath{\response^{(#1)}}\xspace}
\newcommand{\precsteps}[1]{\ensuremath{\response^{#1}}\xspace}
\newcommand{\completion}{\ensuremath{\mathbf{c}}\xspace}
\newcommand{\program}{\ensuremath{\mathcal{P}}\xspace}
\newcommand{\ytoken}[2]{\ensuremath{y^{(#1)}_{#2}}\xspace}

\newcommand{\LM}[1]{\ensuremath{p_{#1}}\xspace}
\newcommand{\RM}[1]{\ensuremath{R_{#1}}\xspace}

\newcommand{\dataset}[1]{\ensuremath{\mathcal{D}_{#1}}\xspace} % definitions

\definecolor{TaskAColor}{RGB}{31, 119, 180}    % Deep Blue
\definecolor{TaskBColor}{RGB}{214, 39, 40}     % Strong Red
\definecolor{OliveGreen}{RGB}{85, 107, 47}
\definecolor{Goldenrod}{RGB}{218, 165, 32}
\definecolor{BrickRed}{RGB}{178, 34, 34}
\definecolor{SeaGreen}{RGB}{46, 139, 87}
\definecolor{DarkOrange}{RGB}{255, 140, 0}
\definecolor{Crimson}{RGB}{220, 20, 60}
\colorlet{QwenColor}{SpringGreen!50}
\colorlet{OursColor}{Apricot!50}

\newcolumntype{C}[1]{>{\columncolor{#1}}c}  % Apply color to a column
\icmltitlerunning{Process Supervision-Guided Policy Optimization for Code Generation}

\begin{document}

\twocolumn[
\icmltitle{Process Supervision-Guided Policy Optimization for Code Generation}
% ]
% %Authors and affiliations
% \author[1]{Ning Dai\footnote[1] \;\;}
% \author[2]{Zheng Wu\footnote[1] \;\;}
% \author[2]{Renjie Zheng}
% \author[2]{Wenlei Shi}
% \author[2]{Xing Jin}
% \author[2]{Guanlin Liu}
% \author[2]{Chen Dun}
% \author[1]{Liang Huang}
% \author[2]{Lin Yan}
% %Affiliations
% \author[ ]{%
% Ning Dai\footnote[1] \; $^{2}$, 
% Zheng Wu\footnote[1] \; $^{1}$, 
% Renjie Zheng$^{1}$, 
% Ziyun Wei$^{1}$, 
% Wenlei Shi$^{1}$, \\ % Line break here
% Xing Jin$^{1}$, 
% Guanlin Liu$^{1}$, 
% Chen Dun$^{1}$, 
% Liang Huang$^{2}$, 
% Lin Yan$^{1}$}

% \affil[1]{ByteDance}
% \affil[2]{Oregon State University}
%\affil[3]{Affiliation 3}
%]

% % Emails
% \affil[ ]{\par\texttt{\{dain,liang.huang\}@oregonstate.edu  \{zheng.wu1,renjie.zheng,neil\}@bytedance.com} }
%\affil[ ]{\par\texttt{ \{zheng.wu1,renjie.zheng,neil\}@bytedance.com} }



\icmlsetsymbol{equal}{*}

\begin{icmlauthorlist}
\icmlauthor{Ning Dai}{equal,osu}
\icmlauthor{Zheng Wu}{equal,bd}
\icmlauthor{Renjie Zheng}{bd}
\icmlauthor{Ziyun Wei}{bd}
\icmlauthor{Wenlei Shi}{bd}
\icmlauthor{Xing Jin}{bd}
\icmlauthor{Guanlin Liu}{bd}
\icmlauthor{Chen Dun}{bd}
\icmlauthor{Liang Huang}{osu}
\icmlauthor{Lin Yan}{bd}
\end{icmlauthorlist}

%\icmlaffiliation{yyy}{Department of XXX, University of YYY, Location, Country}
%\icmlaffiliation{comp}{Company Name, Location, Country}
%\icmlaffiliation{sch}{School of ZZZ, Institute of WWW, Location, Country}

%\setcounter{@affiliationcounter}{2}
\icmlaffiliation{bd}{ByteDance Inc.}
\icmlaffiliation{osu}{Oregon State University}

\icmlcorrespondingauthor{}{\texttt{\{dain,liang.huang\}@oregonstate.edu,\\ \{zheng.wu1,renjie.zheng,neil\}@bytedance.com}}
%\icmlcorrespondingauthor{}{\texttt{ \{zheng.wu1,renjie.zheng,neil\}@bytedance.com}}

%\icmlkeywords{Machine Learning, ICML}

\vskip 0.15in
]



\printAffiliationsAndNotice{\icmlEqualContribution} % otherwise use the standard text.

%\vspace{-5pt}
\begin{abstract}
Reinforcement learning (RL) with unit test feedback has enhanced large language models’ (LLMs) code generation, but relies on sparse rewards provided only after complete code evaluation, limiting learning efficiency and incremental improvements. When generated code fails all unit tests, no learning signal is received, hindering progress on complex tasks. To address this, we propose a Process Reward Model (PRM) that delivers dense, line-level feedback on code correctness during generation, mimicking human code refinement and providing immediate guidance. We explore various strategies for training PRMs and integrating them into the RL framework, finding that using PRMs both as dense rewards and for value function initialization significantly boosts performance. 
%Our approach increases our in-house LLM’s pass rate from 28.2\% to 29.8\% on LiveCodeBench and from 31.8\% to 35.8\% on our internal benchmark. 
Our experimental results also highlight the effectiveness of PRMs in enhancing RL-driven code generation, especially for long-horizon scenarios.
\end{abstract}

\vspace{-22pt}
\section{Introduction}
\label{sec:intro}
\vspace{-2pt}
The rapid advancement of large language models (LLMs) has revolutionized code generation, enabling models to achieve near-human performance on programming tasks~\citep{chen2021evaluating, li2022competition, OpenAI:2023}. These models have demonstrated remarkable abilities to generate syntactically correct and functionally viable code snippets, significantly aiding software development processes.
Building upon these successes, recent research has explored the use of reinforcement learning (RL) from unit test feedback to further enhance the code generation capabilities of LLMs~\citep{le2022coderl, shojaee2023ppocoder, Liu+:2023, dou2024stepcoder}. By incorporating unit tests as a reward mechanism, these methods aim to guide LLMs toward generating code that not only compiles but also passes specified test cases, thereby improving overall code reliability and quality.

However, a significant challenge arises from the nature of the reward signals derived from unit tests. These signals are inherently sparse, as they are only received at the end of an episode after the entire code snippet has been generated and evaluated. This delay in feedback impedes learning efficiency and limits the model's ability to make incremental improvements during code generation. When an LLM fails to generate code that passes any unit tests, it receives no meaningful learning signal, making it difficult to learn to solve more complex coding problems.
In contrast, human programmers typically do not rewrite code from scratch when their programs fail unit tests. Instead, they analyze the code to pinpoint and fix errors, leveraging their understanding of programming logic and structure to iteratively improve upon the current version. This process of step-by-step refinement, which involves receiving and acting upon fine-grained feedback, is missing in the current RL training loop for code generation from unit test feedback.

To address this limitation, we propose integrating a Process Reward Model (PRM)~\citep{Lightman+:2023, WangLSXDLCWS24} into the RL training framework for code generation. A PRM provides dense signals by offering line-level feedback that indicates the correctness of each generated line of code. This fine-grained feedback mechanism mimics the human approach to code refinement and has the potential to enhance learning efficiency by providing immediate guidance during code generation.
While the concept of using PRMs is intuitive, training an effective PRM and integrating it into RL training is non-trivial. Challenges include accurately modeling the correctness of partial code snippets and ensuring stable and effective training dynamics when combining PRM-generated signals with traditional RL methods. Although previous research has attempted to incorporate PRMs into LLM RL training~\citep{WangLSXDLCWS24}, these efforts have been limited to the mathematical domain and have not fully explored the complexities involved.

In this work, we conduct a comprehensive analysis of how PRMs can be integrated into RL training for code generation. We explore various strategies for training a robust code PRM and investigate different methods of utilizing it  to improve code generation performance. Based on our experiments, we provide a practical recipe for successfully using PRMs and integrating them into RL training in the context of code generation. Notably, one of our key findings is that using PRMs concurrently as both dense rewards and value function initialization in RL training  improves learning efficiency and leads to a significant performance improvement.
Our contributions can be summarized as follows:
%\vspace{-5pt}
\begin{itemize}[left=5pt]
    \item %\textbf{A novel framework for integrating process supervision into RL training:} 
    We present a practical and effective pipeline that automatically generates process-level supervision data, trains a PRM based on these supervisions, and integrates it into RL training to provide dense feedback signals. To the best of our knowledge, we are the first to demonstrate that PRMs can benefit RL from unit test feedback in code generation.
    \item %\textbf{Comprehensive analysis of PRM integration strategies:} 
    We systematically investigate how to best integrate PRMs into RL training, exploring different strategies for training high-quality code PRMs and utilizing them to enhance code generation. Our findings are distilled into a practical guideline for effectively incorporating PRMs in RL for code generation.
    \item %\textbf{Significant performance improvements:} 
    Following our proposed methodology, we demonstrate significant improvements in pass rates across HumanEval, MBPP, and LiveCodeBench benchmarks for two baseline LLMs. Additionally, we highlight key insights: (1) the synergy between dense rewards and value initialization maximizes PRM effectiveness, (2) PRMs encourage exploration and improve learning efficiency, and (3) PRMs enhance code generation, particularly in long-horizon scenarios.
\end{itemize}
\vspace{-5pt}

%\vspace{-5pt}
\section{Problem Formalization}
\label{sec:formulation}
%\vspace{-8pt}
In code generation tasks, we define a code generation problem as a sequence of tokens \( \prompt = (\xtoken{1}, \xtoken{2}, \ldots, \xtoken{m}) \), where each \( \xtoken{i} \) denotes the \( i \)-th element or token of the input prompt, which may include problem descriptions.%, code comments, function signatures, etc. 
The primary objective for the model in this context is to process the given input \( \prompt \) and generate a coherent and syntactically correct sequence of code tokens. This sequence is denoted as \( \response = (\step{1}, \step{2}, \ldots, \step{T}) \), where \( T \) represents the total number of code generation steps.
Each individual code generation step, \( \step{t},\ t = 1, 2, \ldots, T \), is composed of a series of tokens \( \ytoken{t}{1}, \ytoken{t}{2}, \ldots, \ytoken{t}{n_t} \), where \( \ytoken{t}{i} \) corresponds to the \( i \)-th token within the \( t \)-th step, and \( n_t \) denotes the number of tokens in this step. 
%The model's challenge is to ensure that each code generation step \( \step{t} \) logically and syntactically builds upon its predecessors, cumulatively leading to a correct and functional piece of code that addresses the initial problem \( \prompt \). 

Typically, a pre-trained language model (LM), denoted as \( \LM{\theta} \), is employed to model the conditional probability distribution of the code generation steps \( \response \), given the code generation problem \( \prompt \), which is mathematically represented as \( \LM{\theta}(\response \mid \prompt) \), parameterized by \( \theta \). The model is optimized through training on a dataset \( \dataset{\prompt \response} \) containing pairs of prompts and their corresponding code solutions. This training process, often referred to as Supervised Fine-Tuning (SFT), involves maximizing the log-likelihood of the dataset.%, as shown in the following equation:
%\begin{equation}\label{eq:SFT}
%    \max_{\theta} \sum_{(\prompt, \response) \in \dataset{\prompt \response}} \log{p_\theta(\response \mid \prompt)}  .
%\end{equation}
\vspace{-5pt}
\subsection{Baseline Method: Reinforcement Learning from Unit Test Feedback}
%\vspace{-5pt}
Code generation tasks can be formulated within a Reinforcement Learning (RL) framework, where code generation is treated as a sequence of decision-making steps. 
% In this view, the code generation process is represented as a trajectory \( \tau = \left\{ (\state{1}, \action{1}), (\state{2}, \action{2}), \ldots, (\state{T}, \action{T}) \right\}\), consisting of state-action pairs,
%\[
%\tau = \left\{ (\state{1}, \action{1}), (\state{2}, \action{2}), \ldots, (\state{T}, \action{T}) \right\},
%\]
% where each state \( \state{t} \) at step \( t \) reflects the prompt \( \prompt \) and the preceding generated code tokens \( \step{1}, \ldots, \step{t-1} \). Specifically, \( \state{1} = \prompt \), and for \( t > 1 \), \( \state{t} = (\prompt, \step{1}, \ldots, \step{t-1}) \). The action \( \action{t} \) corresponds to generating the current code snippet \( \step{t} \). For clarity, we denote the cumulative sequence of tokens up to but not including step \( t \) as \( \precsteps{<t} \), and up to and including step \( t \) as \( \precsteps{\leq t} \).
Once the model has undergone SFT, 
%as expressed in Eq.~\ref{eq:SFT}, 
the RL phase is employed to refine the model’s ability to generate functionally correct code using feedback from unit tests~\citep{Liu+:2023}. Unit test feedback is derived by executing the generated program on predefined test cases. The feedback serves as a signal for learning and can be transformed into a reward. A simple reward function based on the outcome of the unit tests could be defined as follows:
%\begin{small}
\[
\RM{\texttt{UT}}(\prompt, \response) = 
\begin{cases} 
1, & \text{if the program \response passes all unit tests} \\
0, & \text{otherwise}
\end{cases}
\]
%\end{small}
\vspace{-15pt}

This binary reward formulation encourages the model to generate programs that can successfully pass all unit test cases.
Given a collection of unlabeled code generation prompts \( \dataset{\prompt} \), the model \( \LM{\theta} \) is optimized to maximize the expected reward over all possible code generation trajectories. 
%The optimization objective is defined as:
%\[
%\max_{\theta} \sum_{\prompt \in \dataset{\prompt}} \mathbb{E}_{\response  \sim \LM{\theta}(\response \mid \prompt)}\left [ %\RM{\texttt{UT}}(\prompt, \response) \right ]    ,
%\]
%This is typically optimized using policy optimization algorithms, such as Proximal Policy Optimization (PPO) \citep{Schulman+:2017}. 
% which updates the model parameters to increase the likelihood of generating code that passes all unit tests.

%However, this approach presents a significant challenge due to the sparsity of the reward signal. The binary reward function \( \RM{\texttt{UT}} \) provides feedback only at the end of the code generation process, indicating whether the entire program passes all unit tests or not. Consequently, during training, the model receives no gradient information on which parts of the generated code contributed to success or failure. This lack of intermediate feedback makes it difficult for the policy to learn effectively, especially when the probability of generating a program that passes all unit tests is low in the initial stages of training. The sparse reward signal can lead to slow convergence and suboptimal policies, as the model may struggle to discover successful code generation trajectories through random exploration alone.

\section{Process Supervision-Guided \\ \quad Policy Optimization}
\label{sec:method}
%\begin{wrapfigure}{r}{0.98\textwidth} 
\begin{figure*}[!ht]
    \centering
    \includegraphics[width=.85\textwidth]{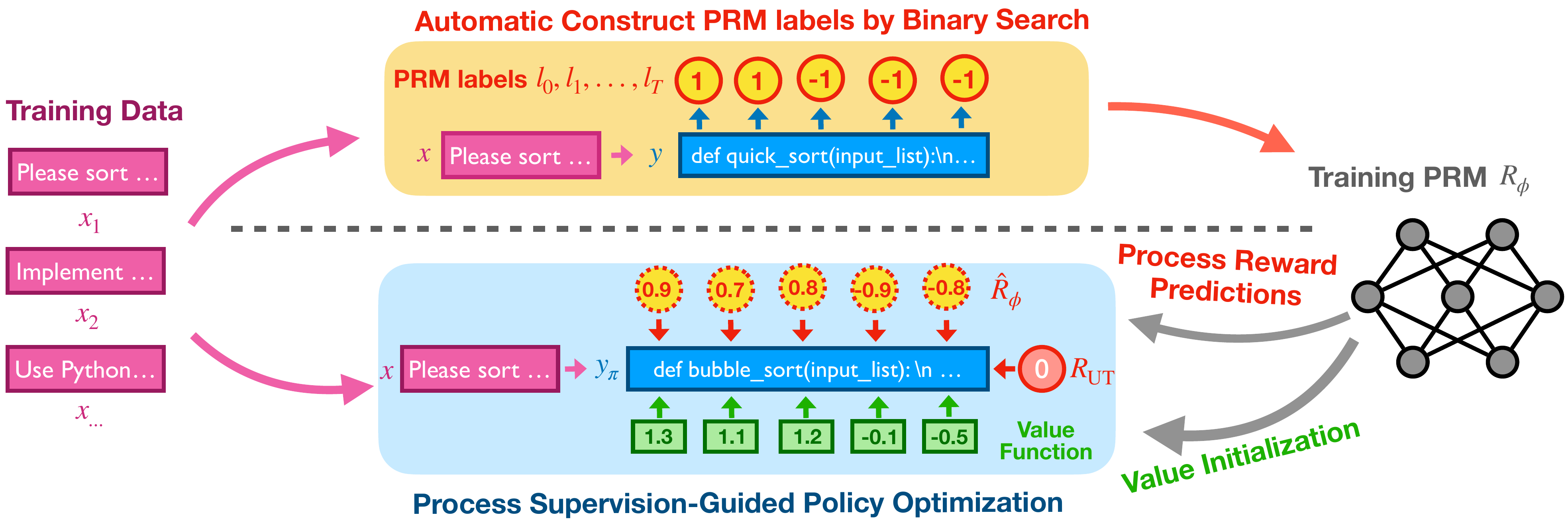}  
    \caption{Overview of our method. Our approach consists of two main components: (1) a binary search-based method for automating PRM training data labeling, used to train a code PRM; and (2) the integration of PRM into RL training, where it serves as (a) the initialization for the value model and (b) an evaluator assessing the correctness of each line of code, providing dense reward signals.}
    \label{fig:intro}
\end{figure*}
%\end{wrapfigure}
%\vspace{-5pt}

While the Reinforcement Learning from Unit Test Feedback (RLTF) offers a framework for improving code generation models, it suffers from significant limitations due to the sparsity of its reward signal. The binary nature of unit test feedback—indicating only whether the entire program passes or fails—provides no guidance on which specific parts of the code contributed to the outcome. This lack of intermediate feedback makes it challenging for the model to identify and correct errors during training, leading to slow convergence and suboptimal performance.
In contrast, human programmers iteratively develop and refine their code. When a program fails to pass unit tests, they do not typically rewrite it from scratch. Instead, they analyze the code to pinpoint and fix errors, leveraging their understanding of programming logic and structure. This process of step-by-step refinement is crucial for efficient problem-solving.

Motivated by this observation, we propose \textbf{Process Supervision-Guided Policy Optimization (PSGPO)}, a method that integrates fine-grained feedback into the RL framework. 
Figure~\ref{fig:intro} illustrates the overview of our approach. 
By providing intermediate rewards that assess the correctness of partial code sequences, our approach guides the model more effectively toward generating functionally correct programs. This is achieved using a Process Reward Model (PRM)~\citep{Lightman+:2023}, which serves as 1) the initialization for the value model and 2) an evaluator for each code generation step, providing dense reward signals that address the limitations of end-of-trajectory rewards.

\subsection{Process Supervision via Process Reward Models}
\label{sec:method:prm}
% \zwu{Emphasize our PRM is defined at the line-level?} 
Our method introduces a PRM to assess the correctness of each \textbf{line} of the code during the generation process. The PRM serves as an oracle that provides intermediate rewards based on the potential of the current code prefix to be extended into a correct program. By offering intermediate feedback, the PRM helps the model identify and reinforce beneficial code generation patterns while discouraging those that introduce errors. This fine-grained feedback mirrors the human approach to coding, where programmers continuously evaluate and adjust their code.

%\newpage
\vspace{-3pt}
\subsubsection{Data Collection}
\label{sec:method:prm:data_collection}

To effectively train the PRM, we require a dataset that provides fine-grained annotations indicating the correctness of partial code sequences. However, manually annotating the correctness of each line of code generated by the model is costly and not scalable. Instead, we employ an automated approach inspired by techniques used in recent works~\citep{WangLSXDLCWS24, wang2024multi, Luo+:2024}. Our method leverages the model's own capabilities to generate completions for partial code prefixes and uses automated testing to assess their correctness.
The key idea is to determine whether a partial code prefix can be extended—by any means—into a complete program that passes all unit tests. If so, we consider the prefix as potentially correct; otherwise, it is labeled as incorrect.

\begin{figure}[h]
    \centering
    \begin{tikzpicture}[scale=0.67]
        % Draw steps
        \foreach \i in {1,...,3} {
            \node[draw, fill=green!90!black, rectangle, minimum width=0.8cm, minimum height=0.8cm] (s\i) at (\i*2,0) {$\step{\i}$};
        }
        \foreach \i in {4,...,5} {
            \node[draw, fill=red!90!black, rectangle, minimum width=0.8cm, minimum height=0.8cm] (s\i) at (\i*2,0) {$\step{\i}$};
        }
        % Draw arrows
        \foreach \i in {1,...,4} {
            \draw[->] (s\i) -- (s\the\numexpr\i+1\relax);
        }
        % Label initial search interval
        \draw[thick, decorate, decoration={brace, amplitude=5pt}] (1,2) -- (10.5,2) node[midway, yshift=10pt]{Initial Interval $[1,5]$};
        % First midpoint
        \draw[green!60!black, thick] (6,1) -- (6,-1.3);
        \node[green!60!black] at (6,1.4) {$m=3$};
        % Accepted prefix
        \node[green!60!black] at (3, -1.5) {Accepted Prefix};
        \draw[green!60!black, thick, ->] (4.5, -1.2) -- (s3.south);
        % Update interval
        \draw[thick, decorate, decoration={brace, amplitude=5pt}] (7,1.2) -- (10.5,1.2) node[midway, yshift=10pt, xshift=5pt]{New Interval $[4,5]$};
        % Second midpoint
        \draw[red, thick] (8,0.8) -- (8,-1.3);
        \node[red] at (8,1) {$m=4$};
        % Rejected prefix
        \node[red] at (8, -1.5) {Rejected Prefix};
        \draw[red, thick, ->] (8, -1.3) -- (s4.south);
    \end{tikzpicture}
    \caption{Binary search over code steps at line level to label prefixes. The first midpoint at \( m=3 \) is accepted, so the search interval moves to \( [4,5] \). The next midpoint at \( m=4 \) is rejected, indicating errors occur after step 3.}
    \label{fig:binary_search}
\end{figure}

Given a prompt \( \prompt \), we generate a complete code response \( \response = (\step{1}, \step{2}, \ldots, \step{T}) \) using the current policy \( \LM{\theta} \). Our goal is to determine the correctness of each partial code prefix \( \precsteps{\leq t} \) for \( t = 1, 2, \ldots, T \). To achieve this, we employ a \textit{best-of-\( K \)} sampling strategy to approximate an oracle capable of completing partial code prefixes.
For each partial code prefix \( \precsteps{\leq t} \), we generate \( K \) potential completions \( \{\completion_k\}_{k=1}^K \) using the current policy. We then form full programs \( \program_k = (\precsteps{\leq t}, \completion_k) \) and execute them against the unit tests \( \mathcal{U} \). If any of these programs pass all unit tests, we label the partial code prefix as \textbf{correct}; otherwise, it is labeled as \textbf{incorrect}.
To efficiently identify the transition point where errors occur, we employ a binary search over the code generation steps~\citep{Luo+:2024}, which is formalized in Algorithm~\ref{alg:binary_search}.
For example, consider a code response divided into five steps (\( T = 5 \)), as shown in Figure~\ref{fig:binary_search}. The partial prefix up to \( \step{3} \) can be completed to pass all unit tests, so it is labeled as correct. The prefix up to \( \step{4} \) cannot, meaning steps beyond \( \step{3} \) are labeled as incorrect.
%\paragraph{Label Assignment}
For each partial code prefix \( \precsteps{\leq m} \), the label \( l_m \) is assigned based on the outcome of the completion attempts:
\vspace{-3pt}
\begin{equation}
\label{eq:prm_label}
    l_m = 
    \begin{cases}
        +1, & \text{if any } \program_k \text{ passes all unit tests} \\
        -1, & \text{otherwise}
    \end{cases}
\end{equation}
which indicate whether the prefix is potentially correct (can be completed to a correct program) or incorrect (contains unrecoverable errors).

\begin{algorithm}[tb]
    \caption{Binary Search for Labeling Code Prefixes}
    \label{alg:binary_search}
    \begin{algorithmic}[1]
        \REQUIRE Prompt $ \prompt $, response $ \response = (\step{1}, \ldots, \step{T}) $, policy $ \LM{\theta} $, unit tests $ \mathcal{U} $, number of completions $ K $
        \ENSURE Labels $ l_t $ for each prefix $ \precsteps{\leq t} $
        \STATE Initialize lower bound $ L \gets 1 $, upper bound $ R \gets T $, failure point $ F \gets T + 1 $
        \WHILE{$ L \leq R $}
            \STATE Compute midpoint $ m \gets \left\lfloor (L + R) / 2 \right\rfloor $ %\dfrac{L + R}{2}
            \STATE Set success flag $ S \gets \text{False} $
            \FOR{$ k = 1 $ to $ K $}
                \STATE Generate completion $ \completion_k \sim \LM{\theta}(\cdot \mid \precsteps{\leq m}) $
                \STATE Form full program $ \program_k \gets (\precsteps{\leq m}, \completion_k) $
                \STATE Execute $ \program_k $ with unit tests $ \mathcal{U} $
                \IF{$ \program_k $ passes all unit tests}
                    \STATE Set $ S \gets \texttt{True}; $ \;\textbf{break}
                \ENDIF
            \ENDFOR
            \STATE $ L \gets \text{if }\; S = \texttt{True} \;\text{ then }\; m + 1 \text{ else } L $
            \STATE $ F \gets \text{if }\; S = \texttt{True} \;\text{ then }\; F \; \text{ else } m $
            \STATE $ R \gets \text{if }\; S = \texttt{True} \;\text{ then }\; R \;\text{ else } m-1 $
        \ENDWHILE
    \FOR{$ t = 1 $ to $ T $}
        \STATE $ l_t \gets \text{if } t < F \text{ then } +1 \text{ else } -1 $
    \ENDFOR
    \end{algorithmic}
\end{algorithm}

%\begin{figure}
% \begin{algorithm}[tb]
%     \caption{Binary Search for Labeling Partial Code Prefixes}
%     %\vspace{10pt} % Vertical space between algorithm and figure
%     \label{alg:binary_search}
%     %\begin{algorithmic}
%         \KwIn{Prompt $ \prompt $, response $ \response = (\step{1}, \ldots, \step{T}) $, policy $ \LM{\theta} $, unit tests $ \mathcal{U} $, number of completions $ K $}
%         \KwOut{Labels $ l_t $ for each prefix $ \precsteps{\leq t} $}
%         Initialize lower bound $ L \gets 1 $, upper bound $ R \gets T $, failure point $ F \gets T + 1 $\;
%         \While{$ L \leq R $}{
%             Compute midpoint $ m \gets \left\lfloor \dfrac{L + R}{2} \right\rfloor $\;
%             Set success flag $ S \gets \text{False} $\;
%             \For{$ k = 1 $ to $ K $}{
%                 Generate completion $ \completion_k \sim \LM{\theta}(\cdot \mid \precsteps{\leq m}) $\;
%                 Form full program $ \program_k \gets (\precsteps{\leq m}, \completion_k) $\;
%                 Execute $ \program_k $ with unit tests $ \mathcal{U} $\;
%                 \If{$ \program_k $ passes all unit tests}{Set $ S \gets \text{True} $; \textbf{break};}
%             }
%             \lIf{$ S = \text{True} $}{
%                 $ L \gets m + 1 $
%             }
%             \lElse{
%                  $ F \gets m $, $ R \gets m - 1 $
%             }
%         }
%         \For{$ t = 1 $ to $ T $}{
%             \lIf{$ t < F $}{$ l_t \gets +1 $}
%             \lElse{$ l_t \gets -1 $}
%         }
%     %\end{algorithmic}
% \end{algorithm}
%\end{figure}

\vspace{-3pt}
\subsubsection{Training}
\label{sec:method:prm:prm_training}

Using the collected data \( \{(\prompt, \precsteps{\leq m}, l_m)\} \), we train the PRM \( \RM{\phi} \) to predict the correctness of partial code prefixes. The PRM learns to assign higher rewards to prefixes labeled as correct and lower rewards to those labeled as incorrect. The training objective, i.e., Mean Squared Error (MSE), minimizes the discrepancy between the PRM's predictions and the assigned labels:
\vspace{-5pt}
\begin{small}
\begin{equation}\label{eq:prm_obj}
    \min_{\phi} \sum_{(\prompt, \precsteps{\leq m})} \left( \RM{\phi}(\prompt, \precsteps{\leq m}) - l_m \right)^2
\end{equation}
\end{small}

\vspace{-10pt}
This regression formulation allows the PRM to estimate the likelihood that a given prefix can be successfully completed. 
Notably, 
%aside from employing Mean Squared Error (MSE) loss, 
we also experimented with Cross-Entropy loss and empirically found that MSE loss yielded better performance.% in our case. 
\vspace{-5pt}
\subsection{Integrating PRM into RL Training}
\label{sec:method:rl}
Given a learned PRM, we aim to identify best practices for enhancing code generation during RL training. While prior work has used PRMs to verify intermediate steps in mathematical tasks~\citep{Lightman+:2023, WangLSXDLCWS24, jiao24learning, wang2024multi, Luo+:2024}, their potential for guiding code generation remains largely unexplored. In mathematical domains, LLMs may generate correct answers with faulty reasoning~\citep{Lightman+:2023}, making intermediate verification essential. However, in code generation, problems are typically accompanied by multiple unit tests, making it improbable for incorrect code to pass all tests. As a result, the emphasis on intermediate verification is less applicable. 
Instead, we propose leveraging PRMs as auxiliary sources of dense signals to facilitate better exploration during RL training. While preliminary attempts have been made to incorporate PRMs into RL training~\citep{WangLSXDLCWS24}, these efforts are limited and warrant a more thorough investigation. We explore the following methods to integrate PRMs effectively:
\vspace{-10pt}
\paragraph{PRM as Dense Rewards.} 
Similar to \citet{WangLSXDLCWS24}, we use PRMs to provide step-level reward signals that guide more efficient policy exploration during RL training. By rating the correctness of each line in the code response, the PRM supplies “dense” rewards that encourage the policy to explore more promising code paths.%, leading to improved performance.
\vspace{-8pt}
\paragraph{PRM as Value Initialization. } The PRM’s method of annotating code, by fixing a prefix  \precsteps{\leq t}  and rolling out the policy to sample correct responses, can be viewed as a “hard” value estimation of  \precsteps{\leq t} . We hypothesize that the PRM’s capability to provide line-level feedback could serve as a useful inductive bias for initializing the value function in RL algorithms, which can ease the credit assignment problem by offering a more informed starting point. %, replacing the common practice of using a learned reward model \cite{Ouyang+:2022}. 
%\vspace{-8pt}
%\paragraph{PRM as Both Dense Rewards and Value Initialization.} 
%To fully capitalize on the advantages of PRMs, we combine both approaches. By using PRMs for dense rewards and as an initialization for the value model, we aim to enhance RL training through improved exploration and more effective credit assignment.

%\vspace{-5pt}

\section{Experimental Results}
\label{sec:exp}

\vspace{-5pt}
\subsection{Experimental Setup}
\label{sec:exp:setup}
\vspace{-5pt}
\paragraph{Datasets and Evaluation.} 
We utilize in-house datasets to train our model for code generation. 
Specifically, the training set, $\mathcal{D}_{\text{train}}$, is a comprehensive Reinforcement Learning with Human Feedback (RLHF) dataset that includes, as a subset, approximately $30,000$ diverse coding problems. Each of these problems is paired with unit tests designed to validate the functional correctness of the generated code. 
For evaluation, we use three widely adopted benchmarks: \textbf{HumanEval}~\citep{HumanEval}, which contains 164 hand-crafted programming problems; \textbf{MBPP}~\citep{MBPP}, a dataset of 974 crowd-sourced Python programming problems (where problems with IDs 11–510 are used for evaluation); and \textbf{LiveCodeBench}~\citep{jain2024livecodebench}, a comprehensive benchmark designed to assess the code generation capabilities of LLMs. Among the various releases of LiveCodeBench, we use \textbf{LiveCodeBench v3}, which includes 612 coding tasks collected between May 2023 and July 2024.
In our experiments, we restricted LLMs to a single-turn chat completion setting. For each coding problem, we directly input the problem to the model without using few-shot prompting. We generate 10 candidate responses for each problem, using a temperature of 0.2, nucleus sampling with top-$p \text{=} 0.95$, and top-k sampling with $k \text{=} 128$, following common practice. We adopt Pass@1 as the evaluation metric, in line with previous work~\citep{Kulal+:2019, chen2021evaluating, jain2024livecodebench}.

\vspace{-5pt}
\paragraph{Base Models.} 
In our experiments, we employ two different base models, \texttt{Qwen2.5-7B} and \texttt{Doubao-Lite}, to evaluate the effectiveness of our proposed method.
\texttt{Qwen2.5-7B}~\citep{qwen2.5} is a publicly released causal language model from the Qwen series~\citep{qwen2}, recognized for its strong performance across diverse domains.
\texttt{Doubao-Lite} is an in-house model of comparable size and capability to \texttt{Qwen2.5-7B} but utilizes a different network architecture.

\paragraph{SFT and RL Baseline.}
Initially, both \texttt{Qwen2.5-7B} and \texttt{Doubao-Lite} are fine-tuned on our Supervised Fine-Tuning (SFT) dataset, yielding \texttt{Qwen2.5-7B-SFT} and \texttt{Doubao-Lite-SFT}, which serve as the starting points for the subsequent RLHF training phase.
We then further optimize these SFT models ($\pi_\mathrm{ref}$) on the RLHF dataset $\mathcal{D}_{\text{train}}$ using Proximal Policy Optimization (PPO)~\citep{Schulman+:2017}, resulting in the RL models \texttt{Qwen2.5-7B-RL} and \texttt{Doubao-Lite-RL} ($\pi_{\theta}$). In our setup, two types of Outcome Reward Models (ORMs) are employed as the objective functions for RL training. For non-coding prompts, we use a general reward model, $\RM{\texttt{general}}(\prompt, \response)$, derived from preference learning on a human-annotated dataset~\citep{Ouyang+:2022}. For coding prompts, the ORM is defined as a binary indicator of whether the response passes all unit tests, \RM{\texttt{UT}}(\prompt, \response). Following ~\cite{Ouyang+:2022}, RLHF optimization objective is defined as:
\vspace{-5pt}
\[
\max_{\theta} \sum_{\prompt \in \dataset{train}} \mathbb{E}_{\response  \sim \pi_{\theta}(\response \mid \prompt)}\left [ R(\prompt, \response) - \beta \mathrm{KL}(\pi_\theta \parallel \pi_\mathrm{ref}) \right ]    ,
\]
with $R(\prompt, \response) = \RM{\texttt{general}}(\prompt, \response)$ for non-coding prompts and $R(\prompt, \response) = \RM{\texttt{UT}}(\prompt, \response)$ for coding prompts.

\vspace{-5pt}
\paragraph{PRM Training.}
To ensure that the PRM training data effectively covers the state space the language model may encounter during the next RL training phase, we sample policy models from various stages of the RL baseline training. Specifically, we select 4 checkpoints evenly spaced throughout the RL baseline model's training process. For each checkpoint, we sample $n$ responses for each coding prompt in the training dataset $\mathcal{D}_{\text{train}}$. For each sampled response, we apply the binary search labeling procedure described in Algorithm~\ref{alg:binary_search}, using $K = 20$ completions for each partial code prefix. 
The data collected from all checkpoints is then aggregated into a PRM training set, denoted as $\mathcal{D}_{\text{PRM}}$. We initialize the PRM with the value model from the RL baseline and fine-tune it on the aggregated dataset, $\mathcal{D}_{\text{PRM}}$, using the objective function defined in Eq.~\eqref{eq:prm_obj}.

\vspace{-5pt}
\paragraph{Integrating PRM into RL.}
As described in Section~\ref{sec:method:rl}, we explore two methods for integrating the Process Reward Model (PRM) into RL training: 1) using PRM as a source of dense reward signals (\textbf{DenseReward}) and 2) initializing the value function in PPO with PRM (\textbf{ValueInit}). In the \textbf{DenseReward} approach, PRM assigns additional reward signals at each end-of-line token (\texttt{\textbackslash n}) in the code response for coding prompts. Thus, the RL optimization objective for coding prompts is modified to the weighted sum of \RM{\texttt{UT}} and \RM{\texttt{PRM}}, as defined below:
\begin{equation}
\label{eq:rlhf_obj}
    \begin{aligned}
    \max_{\theta} \sum_{\prompt \in \dataset{train}} \mathbb{E}_{\response  \sim \pi_{\theta}(\response \mid \prompt)} [ & \RM{\texttt{UT}}(\prompt, \response) + \lambda \RM{\texttt{PRM}}(\prompt, \response) \\
    & \qquad - \beta \mathrm{KL}(\pi_\theta \parallel \pi_\mathrm{ref})  ]    ,
    \end{aligned}
\end{equation}
where $\lambda$ controls the relative importance of PRM in shaping the reward. Specifically, we set $\lambda = 0.25$ when the code response does not pass all unit tests, i.e., $\RM{\texttt{UT}}(\prompt, \response) = 0$, and $\lambda = 0.025$ when the response passes all unit tests, i.e., $\RM{\texttt{UT}}(\prompt, \response) = 1$. The intuition behind this reward shaping is to leverage PRM to provide informative signals when the RL policy fails to generate a valid solution, while minimizing the risk of PRM over-optimization~\citep{rafailov2024scaling, skalse2022defining} once a correct solution is found. Our empirical results indicate that this reward shaping strategy performs effectively in our experimental setting. 
In the \textbf{ValueInit} setting, PRM is simply used to initialize of the value function in PPO. 
Notably, these two approaches-\textbf{DenseReward} and \textbf{ValueInit}—are complementary and can be applied concurrently.

\subsection{Key Aspects for Integrating PRM into RL Training}
\label{sec:exp:important_details}
While integrating PRM into RL training might seem straightforward, we found that achieving effective results requires careful attention to several critical factors. In this section, we highlight key implementation details essential for the successful application of PRM in RL training.
\subsubsection{PRM Training: More Data or Better Data?}
\label{sec:exp:important_details:distribution}
Recent research on LLMs highlights that data quality often outweighs quantity~\citep{Gunasekar+:2023, Li+:2023}. We found the same holds true for PRM training data selection. 
%Although automated data annotation allows for the generation of large volumes of PRM training data through model sampling, our experiments showed that increasing data volume can sometimes degrade PRM performance when integrated into RL. In contrast, a smaller, carefully curated subset of the full dataset led to better supervision and improved outcomes.
%For example, when all sampled responses to a given prompt either consistently pass (or fail) unit tests, PRM gains little useful information. In such cases, the model can only learn to predict the correct (or incorrect) label when it encounters the same prompt again, limiting its ability to generalize. We explored various strategies for selecting and filtering data, as detailed in Section~\ref{sec:exp:main_results}.
While automated annotation enables large-scale data generation via model sampling, our experiments showed that increasing volume can sometimes degrade PRM performance in RL. Instead, a smaller, well-curated subset provided better supervision and improved outcomes.
For instance, when all sampled responses to a prompt consistently pass or fail unit tests, PRM learns little beyond memorization, limiting generalization. We explored various selection and filtering strategies to mitigate this, as detailed in Section~\ref{sec:exp:main_results}.

\vspace{-5pt}
\subsubsection{RL Training: Alleviating PRM Hacking} 
\label{sec:exp:important_details:hack}
Reward model hacking~\citep{skalse2022defining} is a well-known issue in RLHF training, where the policy learns to exploit flaws in the reward model to achieve high rewards without genuinely improving the quality of response. 
Similarly, we observed that PRM is also susceptible to such exploitation. 
Here we discuss two key practical strategies to mitigate the risk of PRM hacking and ensure the reward signals remain aligned with the intended task objectives.

\paragraph{PRM Reward Length Normalization.} 
As described in Section~\ref{sec:exp:setup}, PRM provides dense rewards by assigning line-level signals at end-of-line tokens. However, directly using PRM predictions, $\RM{\phi}$, as the reward signal $\RM{\mathrm{PRM}}$ in~\ref{eq:rlhf_obj} allows exploitation: the policy can generate excessive lines with positive PRM rewards, artificially inflating the total reward. To prevent this, we apply length normalization. Given a prompt \prompt and a response \response with $T$ lines, \response = (\step{1}, \step{2}, \ldots, \step{T}), the PRM dense reward at line $m$ is:
\[
\RM{\mathrm{PRM}}(\response^{(m)}) = \frac{1}{T} \cdot \RM{\phi}(\prompt, \precsteps{\leq m}),
\]
which ensures rewards remain bounded in $[-1,1]$, preventing the policy from gaining an advantage by generating excessively long responses.

\vspace{-5pt}
\paragraph{Neutral Labeling in PRM Training.}
While length normalization curbs reward inflation, models can still exploit PRM by generating excessive comments, which are easier to write than correct code. To address this, we introduce a neutral label in PRM annotation, as an extention of Eq.~\eqref{eq:prm_label}:
\vspace{-3pt}
\[
l_m = 
    \begin{cases}
        +1, & \text{if any } \program_k \text{ passes all unit tests} \\ 
        0, & \text{if the line is a comment} \\
        -1, & \text{otherwise}
    \end{cases}
\]
\vspace{-10pt}

By assigning a neutral label (0) to comments, we eliminate the incentive to generate unnecessary comments, ensuring PRM rewards only meaningful code contributions.

\newcommand{\cmark}{\checkmark} % Checkmark symbol
\newcommand{\xmark}{\texttimes} % X mark symbol
\begin{table*}[!ht]
    \centering
    \vspace{-0.1in}
    \caption{Model performance comparison (Pass@1) on HumanEval, MBPP, and LiveCodeBench datasets. The first section presents the results of public models, alongside the models used in our experiments (\colorbox{QwenColor}{Qwen2.5-7B} and \colorbox{OursColor}{Doubao-Lite} series). The second and third sections detail the performance of RL models under different configurations of PRM usage. Notably, for both models, the best overall performance is achieved when PRM is applied for both Dense Reward and Value Initialization (ValueInit). }
    \vspace{0.1in}
    \label{tab:performance}
    \resizebox{.95\textwidth}{!}{
    \renewcommand{\arraystretch}{0.85}
    \begin{tabular}{lcccccccc}
        \toprule
        \multirow{3}{*}{\textbf{Model}} & \multicolumn{2}{c}{\textbf{Setting}} & \multicolumn{6}{c}{\textbf{Dataset}} \\ 
        \cmidrule(lr){2-3} \cmidrule(lr){4-9}
        & \textbf{Dense} & \textbf{Value} & \multirow{2}{*}{\textbf{HumanEval}} & \multirow{2}{*}{\textbf{MBPP}} & \multicolumn{4}{c}{\textbf{LiveCodeBench}} \\ % & \multicolumn{3}{c}{\textbf{InHouseBench}} \\ 
        %\cmidrule(lr){4-7} \cmidrule(lr){8-10}
        \cmidrule(lr){6-9}
        & \bf{Reward} & \bf{Init.} &  &   & Easy & Medium & Hard & Overall \\ %& Contest & NL2Alg & Overall \\ 
        \midrule
        %-0718
        GPT-4o-mini & \cellcolor{white} - & \cellcolor{white} - & \cellcolor{white} 87.2 & \cellcolor{white} 71.8 & \cellcolor{white} 81.9 & \cellcolor{white} 27.2 & \cellcolor{white} 3.6 & \cellcolor{white} 40.7 \\ %& 43.8 & 68.4 & 51.4 \\ 
        %Ins
        Qwen2-72B & \cellcolor{white} - & \cellcolor{white} - & \cellcolor{white} 64.6 & \cellcolor{white} 76.9 & \cellcolor{white} 65.0 & \cellcolor{white} 21.3 & \cellcolor{white} 2.8 & \cellcolor{white} 32.2 \\ %& 14.8 & 51.3 & 26.1 \\ 
        %Gemini-Flash-1.5-May
        %Gemini-Flash-1.5 & - & - & 74.3 & - & 67.7 & 13.1 & 1.9 & 29.6 \\ %& - & - & - \\
        %Ins
        DeepseekCoder-33B & \cellcolor{white} - & \cellcolor{white} - & \cellcolor{white} 79.3 & \cellcolor{white} 70.0 & \cellcolor{white} 60.8 & \cellcolor{white} 14.8 & \cellcolor{white} 1.2 & \cellcolor{white} 27.7 \\ %& 10.3 & 50.3 & 22.7 \\  
        %\midrule
         Qwen2.5-7B-SFT & \cellcolor{QwenColor} - & \cellcolor{QwenColor} - & \cellcolor{QwenColor} 67.8 & \cellcolor{QwenColor} 58.1 & \cellcolor{QwenColor} 50.7 & \cellcolor{QwenColor} 16.5 & \cellcolor{QwenColor} 0.9 & \cellcolor{QwenColor} 24.9 \\ %& 12.3 & 35.4 & 19.4 \\ 
        Doubao-Lite-SFT & \cellcolor{OursColor} - & \cellcolor{OursColor} - & \cellcolor{OursColor} 59.3 & \cellcolor{OursColor} 59.9 & \cellcolor{OursColor} 55.3 & \cellcolor{OursColor} 9.3 & \cellcolor{OursColor} 0.3 & \cellcolor{OursColor} 23.5 \\ %& 10.4 & 41.4 & 20.0 \\ 
        \specialrule{0.5pt}{1pt}{1pt}
         \multirow{4}{*}{Qwen2.5-7B-RL} & \cellcolor{QwenColor} \xmark & \cellcolor{QwenColor} \xmark & \cellcolor{QwenColor} 73.8 & \cellcolor{QwenColor} 62.4 & \cellcolor{QwenColor} 60.9 & \cellcolor{QwenColor} 13.7 & \cellcolor{QwenColor} 1.4 & \cellcolor{QwenColor} 27.5 \\ %& 26.0 & 45.6 & 32.0 \\ 
         & \cellcolor{QwenColor} \xmark & \cellcolor{QwenColor} \cmark & \cellcolor{QwenColor} 75.4 & \cellcolor{QwenColor} 63.1 & \cellcolor{QwenColor} 62.8 & \cellcolor{QwenColor} \textbf{17.1} & \cellcolor{QwenColor} \textbf{1.7} & \cellcolor{QwenColor} 29.6 \\ %& \textbf{30.6} & 46.1 & \textbf{33.6} \\ 
          & \cellcolor{QwenColor} \cmark & \cellcolor{QwenColor} \xmark & \cellcolor{QwenColor} \textbf{76.0} & \cellcolor{QwenColor} 63.4 & \cellcolor{QwenColor} 63.1 & \cellcolor{QwenColor} 14.5 & \cellcolor{QwenColor} 1.1 & \cellcolor{QwenColor} 28.5 \\ %& 27.3 & 47.6 & \textbf{33.6}\\ 
         & \cellcolor{QwenColor} \cmark & \cellcolor{QwenColor} \cmark & \cellcolor{QwenColor} 74.3 & \cellcolor{QwenColor} \textbf{65.4} &  \cellcolor{QwenColor}\textbf{66.3} & \cellcolor{QwenColor} 15.3 & \cellcolor{QwenColor} \textbf{1.7} & \cellcolor{QwenColor} \textbf{30.1} \\ %& 26.1 & \textbf{48.7} & 33.1 \\ 
        %\bottomrule
        \specialrule{0.5pt}{1pt}{1pt}
        \multirow{4}{*}{Doubao-Lite-RL} & \cellcolor{OursColor} \xmark & \cellcolor{OursColor} \xmark & \cellcolor{OursColor} 65.1 & \cellcolor{OursColor} 61.9 & \cellcolor{OursColor} \textbf{70.0} & \cellcolor{OursColor} 7.2 & \cellcolor{OursColor} 1.7 & \cellcolor{OursColor} 28.2 \\ %& 24.4 & 48.7 & 31.8 \\ 
         & \cellcolor{OursColor} \xmark & \cellcolor{OursColor} \cmark & \cellcolor{OursColor} 69.8 & \cellcolor{OursColor} 63.3 & \cellcolor{OursColor} 67.9 & \cellcolor{OursColor} 8.9 & \cellcolor{OursColor} 1.9 & \cellcolor{OursColor} 28.2 \\ %& 25.0 & 45.4 & 31.4 \\ 
         & \cellcolor{OursColor} \cmark & \cellcolor{OursColor} \xmark & \cellcolor{OursColor} 70.0 & \cellcolor{OursColor} 62.1 & \cellcolor{OursColor} 68.5 & \cellcolor{OursColor} 9.9 & \cellcolor{OursColor} \textbf{2.5} & \cellcolor{OursColor} 28.9 \\ %& 25.2 & 48.1 & 32.3 \\ 
         & \cellcolor{OursColor} \cmark & \cellcolor{OursColor} \cmark & \cellcolor{OursColor} \textbf{70.9} & \cellcolor{OursColor} \textbf{63.8} & \cellcolor{OursColor} 69.3 & \cellcolor{OursColor} \textbf{12.0} & \cellcolor{OursColor} 1.6 & \cellcolor{OursColor} \textbf{29.8} \\ %& \textbf{27.9} & \textbf{53.5} & \textbf{35.8} \\ 
        \bottomrule
    \end{tabular}
    }
\end{table*}

%\vspace{-5pt}
\subsection{Main Results and Analysis}
\label{sec:exp:main_results}

\paragraph{Comparing PRM Integration Strategies in RL Training.} 
We evaluate three strategies for incorporating PRM into RL training, as outlined in Section~\ref{sec:exp:setup}: \textit{DenseReward}, \textit{ValueInit}, and a combined approach (\textit{DenseReward \& ValueInit}). Table~\ref{tab:performance} presents the performance of RL models trained using these strategies on HumanEval, MBPP, and LiveCodeBench, alongside SFT and RL baselines. For reference, we also include results from publicly available models such as GPT-4o-mini~\cite{OpenAI:2023}, Qwen2-72B~\cite{bai2023qwen}, and DeepseekCoder-33B~\cite{guo2024deepseekcoderlargelanguagemodel}.

Our results indicate that using PRM as a dense reward signal significantly improves performance over the RL baseline (see the first and third settings for \texttt{Qwen2.5-7B-RL} and \texttt{Doubao-Lite-RL} in Table~\ref{tab:performance}), aligning with findings from~\cite{WangLSXDLCWS24}. The granular feedback provided by PRM facilitates policy exploration by offering continuous corrections at intermediate steps. Additionally, using PRM solely for value function initialization also yields consistent improvements. In \texttt{Qwen2.5-7B-RL}, this setting outperforms the RL baseline across all benchmarks, while in \texttt{Doubao-Lite-RL}, similar gains are observed on HumanEval and MBPP.

Combining PRM for both dense rewards and value initialization yields significant performance improvements. In \texttt{Qwen2.5-7B-RL}, Pass@1 increases from $62.4\%$ to $65.4\%$ on MBPP and from $27.5\%$ to $30.1\%$ on LiveCodeBench. Similarly, in \texttt{Doubao-Lite-RL}, Pass@1 improves from $65.1\%$ to $70.9\%$ on HumanEval, $61.9\%$ to $63.8\%$ on MBPP, and $28.2\%$ to $29.8\%$ on LiveCodeBench. This improvement stems from the complementary roles of PRM: dense rewards facilitate exploration by providing rich intermediate feedback, while value initialization stabilizes training and enhances credit assignment. Together, these mechanisms accelerate convergence toward optimal solutions, driving the observed performance gains.

\vspace{-5pt}
\paragraph{PRM Encourages Exploration and Improves Learning Efficiency.}

\begin{figure}[th]
    \centering
    \includegraphics[width=.95\linewidth]{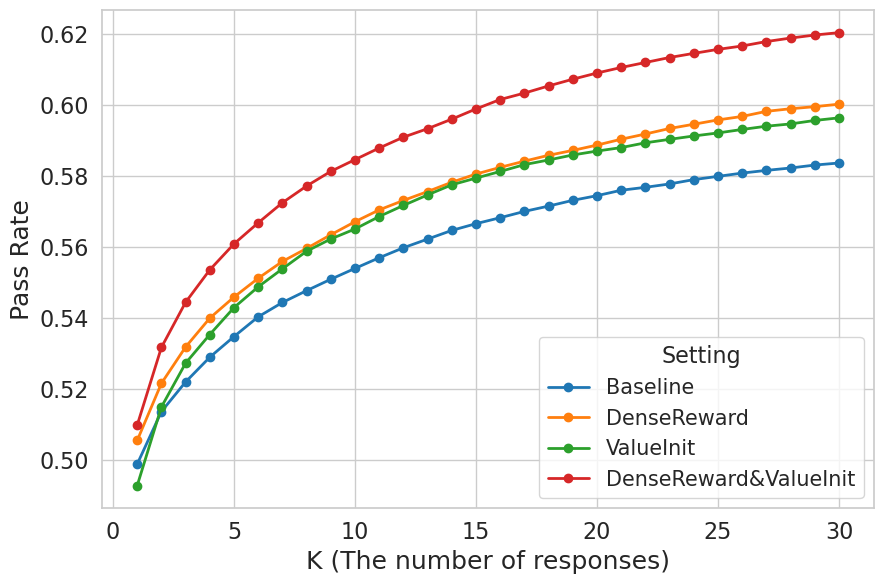}  
    \caption{Best-of-K performance curves for all RL training settings, showing the percentage of problems solved within $K$ generated responses.}
    \label{fig:bon_curve}
\end{figure}

We evaluated the Best-of-K performance for all four training configurations in \texttt{Doubao-Lite-RL} experiments on the training set. Specifically, we assessed all RL models using a decoding configuration with a temperature of 1.0, nucleus sampling (top-$p = 0.95$), and top-k sampling ($k = 128$). For each $K$, we recorded the percentage of problems solved within $K$ generated responses, referred to as the \textit{Pass Rate}.
Figure~\ref{fig:bon_curve} presents the results for $K$ ranging from 1 to 30. Both \textit{DenseReward} and \textit{ValueInit} individually enhance Best-of-K performance compared to the baseline. When combined, they yield the highest improvement, with a nearly 4\% increase in Pass Rate at $K=30$ over the baseline, highlighting the synergy between dense rewards and value initialization.

\vspace{-5pt}

\paragraph{PRM Enhances Code Generation in Long-Horizon Scenarios.}

To understand when PRM provides the greatest benefit, we analyze its impact based on response length. Intuitively, the dense nature of the reward signals provided by PRM is particularly advantageous for long-horizon problems, where intermediate feedback can guide policy exploration more effectively. 

\begin{figure}[th]
%\vspace{-15pt}
    \centering
    \includegraphics[width=.45\textwidth]{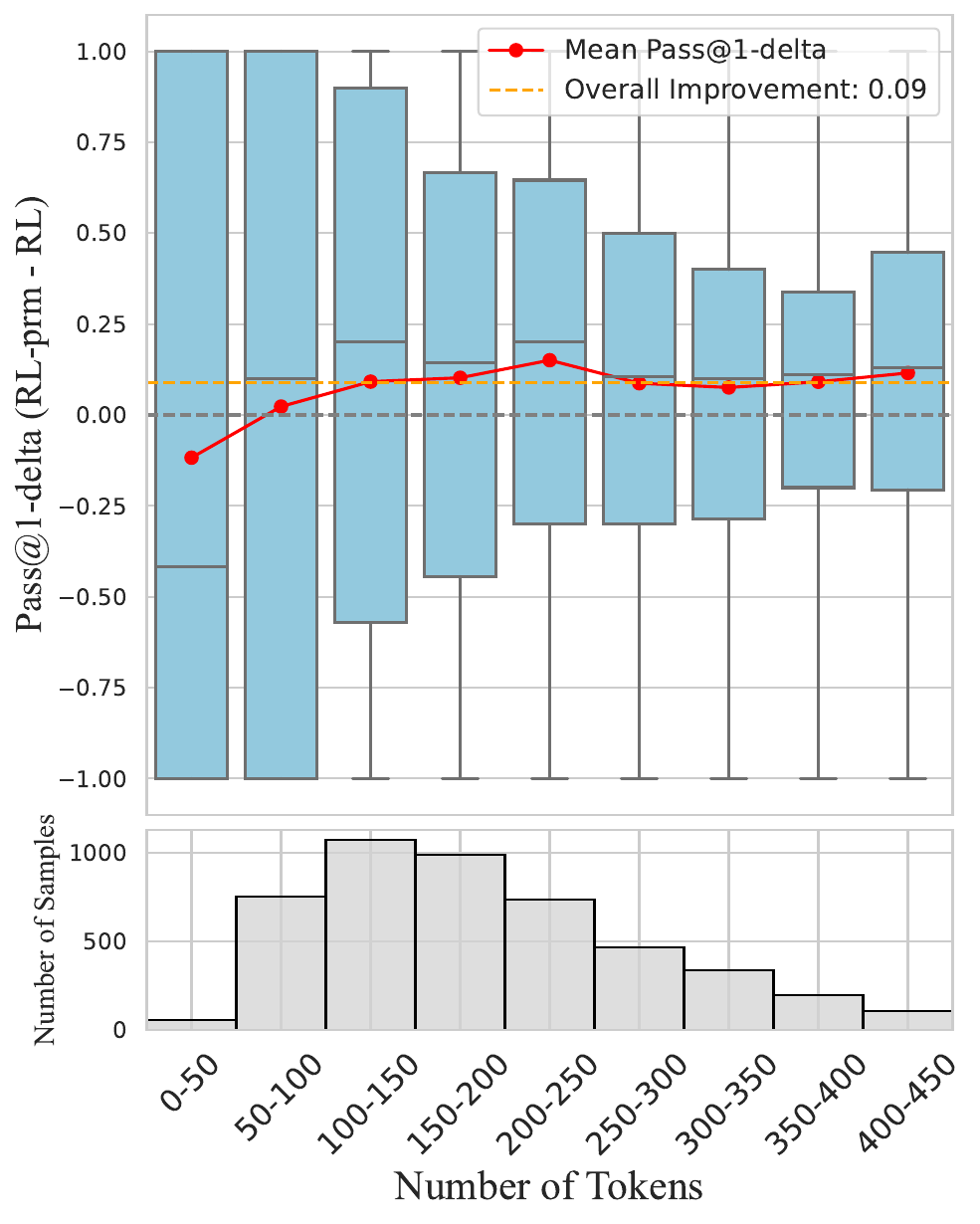}  
    \caption{Pass@1 difference between policies trained with and without PRM across varying response lengths. Policies trained with PRM exhibit consistent improvements over those without PRM for longer-horizon responses (greater than 100 tokens). This demonstrates PRM’s effectiveness in providing intermediate feedback, thereby enabling RL to do more explorations. 
    %during training.
    }
    \label{fig:pass1_delta_vs_num_tokens}
\end{figure}

To validate this, we compared Pass@1 performance of models trained with and without PRM across different response lengths, as shown in Figure~\ref{fig:pass1_delta_vs_num_tokens}. Overall, PRM-trained models achieve a 9\% improvement in Pass@1 over the baseline. Notably, PRM consistently enhances performance for responses exceeding 100 tokens, whereas for shorter responses, its effect is neutral or slightly negative.
We hypothesize that in short-horizon problems, PRM behaves similarly to a biased ORM, offering limited improvements since these problems are already well-explored by the policy. In contrast, for complex, long-horizon problems, PRM provides valuable intermediate signals that help the policy navigate the solution space more effectively, achieving better results with the same optimization compute.

%\vspace{-20pt}
%\end{wrapfigure}

\paragraph{The Importance of PRM Training Data Selection} 
PRM training data can be categorized at two levels: At the \textit{response level}, responses are classified as {\textcolor{SeaGreen}{Correct}} (passes unit tests immediately), {\textcolor{DarkOrange}{Revised}} (initially fails but can find a correct prefix), and {\textcolor{Crimson}{Wrong}} (cannot find any correct prefix by binary search within the given budget). At the \textit{prompt level}, prompts are categorized as \textbf{\textcolor{OliveGreen}{Easy}} (all responses are \textcolor{SeaGreen}{Correct}), \textbf{\textcolor{Goldenrod}{Medium}} (mixed response types), and \textbf{\textcolor{BrickRed}{Hard}} (all responses are \textcolor{Crimson}{Wrong}). We tested the following data selection strategies: \textbf{Full} (use all collected data), \textbf{Remove \textcolor{BrickRed}{Hard}} (exclude \textcolor{BrickRed}{Hard} prompts and their responses), \textbf{\textcolor{Goldenrod}{Medium} Only} (include only prompts with mixed response types), and \textbf{\textcolor{DarkOrange}{Revised} Only} (use only \textcolor{DarkOrange}{Revised} responses). 
We empirically found that \textbf{\textcolor{DarkOrange}{Revised} Only}, which includes the richest process-level correction signals, performs best in our setting.

%\begin{figure}[ht]
%\begin{minipage}[b]{0.4\textwidth}
%\begin{wraptable}{l}{0.45\textwidth}
\begin{table}[ht]
%\vspace{+10pt}
    \centering
    \vspace{-0.1in}
    \caption{Comparison of different PRM data selection strategies on LiveCodeBench (\textbf{LCB}) with \texttt{Doubao-Lite-RL} models.}
    \vspace{0.1in}
    \resizebox{\linewidth}{!}{
    \begin{tabular}{l|cccc}
        \toprule
         & \textbf{Full} & \textbf{Remove \textcolor{BrickRed}{Hard}} & \textbf{\textcolor{Goldenrod}{Medium} Only} & \textcolor{DarkOrange}{Revised} \textbf{Only}  \\ \midrule
        \textbf{LCB} & 26.9   & 27.8 & 26.9 & 29.8 \\ 
        \bottomrule
    \end{tabular}
    }
    \label{tab:asorm}
%\end{wraptable}
%\end{minipage}%
\end{table}

\paragraph{How much data needed to train a PRM that benefits RL training?}

Given that automatic PRM data collection is computationally expensive, we examine how the performance of policies trained with PRM scales with the number of training samples. Figure~\ref{fig:performance} shows how the pass rate of \texttt{Doubao-Lite-RL} models trained with varying amounts of PRM data changes along the average number of responses collected per prompt for PRM data collection, as described in Section~\ref{sec:method:prm:data_collection}. The key finding is that the performance of models trained with PRM improves consistently as the number of PRM training samples increases, highlighting the effectiveness and scalability of our approach.

\begin{figure}[h]
%\vspace{-15pt}
    \centering
    \resizebox{\linewidth}{!}{ % Resize the plot to 48% of text width for a smaller plot
    \begin{tikzpicture}
        \begin{axis}[
            width=1.2\linewidth, % Make the plot wider
            height=0.55\linewidth, % Keep the height smaller to enhance X-to-Y ratio
            xlabel={\parbox{8cm}{Average Number of Collected Responses per Prompt %for PRM Data Collection
            }},
            ylabel={\large Pass@1 ($\%$)},
            xmin=0.25, xmax=8, % Adjusted X-axis range
            ymin=28, ymax=30, % Y-axis remains between 0 and 1
            xmode=log, % Set X-axis to log scale
            log basis x=2, % Log base 2 for X-axis
            grid=both, % Added gridlines
            major grid style={line width=0.5pt,draw=gray!50},
            minor grid style={draw=gray!20},
            xtick={0.25, 0.5, 1, 2, 4, 8}, % Logarithmic X-axis ticks
            ytick={28.0, 28.4, 28.8, 29.2, 29.6, 30.0}, % Y-axis tick positions
            tick align=outside,
            tick label style={font=\footnotesize}, % Smaller font for tick labels
            label style={font=\normalsize}, % Normal font size for axis labels
            legend style={font=\small, at={(0.95,0.05)},anchor=south east}, % Legend inside plot
            mark size=3pt, % Slightly increased marker size for better visibility
            axis line style={thick},
            %smooth, % Smooth curves
            line width=1.2pt % Thicker lines for better clarity
        ]

        % Task A Performance (blue gradient curve)
        \addplot[
            color=blue!70!cyan,
            mark=square*,
            mark options={fill=blue!70!cyan},
            line width=1.2pt
        ] coordinates {
            (0.25, 28.4)
            (1, 28.9)
            (2, 28.9)
            (4, 29.8)
            (8, 30.0)
        };
        \addlegendentry{LiveCodeBench}
        \end{axis}
    \end{tikzpicture}
    } % End of resizebox
    \vspace{-15pt}
    \caption{Pass@1 on LiveCodeBench as the average number of responses per prompt for PRM data collection increases (logarithmic scale). A value of $< 2^0$ indicates that we subsampled prompts from the full dataset, resulting in a smaller prompt set.}
    \label{fig:performance}
%\end{wrapfigure}
%\end{minipage}
\vspace{-10pt}
\end{figure}

\section{Related Works}
\label{sec:related}

\subsection{LLMs for Code Generation}
Recently, large language models (LLMs) have demonstrated impressive capabilities in code generation by pre-training on vast text datasets that include code~\citep{lu2021codexgluemachinelearningbenchmark, christopoulou2022pangucoderprogramsynthesisfunctionlevel, allal2023santacoderdontreachstars, zheng2024codegeexpretrainedmodelcode, Li+:2023}. Additionally, models fine-tuned through supervised fine-tuning (SFT) have achieved competitive results in code generation tasks~\citep{chen2021evaluating, li2023starcodersourceyou, luo2023wizardcoderempoweringcodelarge, rozière2024codellamaopenfoundation, guo2024deepseekcoderlargelanguagemodel}.
Reinforcement Learning (RL) optimizes policies by interacting with an environment and receiving rewards~\citep{williams1992simple}. Recently, RL has been incorporated into LLMs to enhance code generation using unit test feedback~\citep{shojaee2023ppocoder,Liu+:2023,le2022coderl}. CodeRL~\citep{le2022coderl} applies unit test signals as rewards with an actor-critic method, while PPOCoder~\citep{shojaee2023ppocoder} builds on this by using the PPO algorithm. RLTF~\citep{Liu+:2023} improves precision by locating errors, though the reward space remains sparse. Despite progress, RL’s potential to significantly boost code generation in sparse reward environments remains underexplored.

\subsection{Process Reward Models}
Process reward models (PRMs) have garnered significant attention in recent LLM developments, particularly in the mathematical reasoning domain, where they provide verification for intermediate reasoning steps~\citep{Lightman+:2023, WangLSXDLCWS24, jiao24learning, wang2024multi, Luo+:2024}. While some approaches rely on costly and resource-intensive human-annotated process data~\citep{Lightman+:2023}, recent research has focused on automating the collection of process supervision data~\cite{WangLSXDLCWS24, jiao24learning, wang2024multi, Luo+:2024}. Building on these efforts, we similarly automate process supervision but differ in our primary objective. Rather than using PRMs solely as enhanced verifiers compared to Outcome Reward Models (ORMs), we focus on their integration into RL training for code generation. While~\cite{WangLSXDLCWS24} provides preliminary results on PRMs improving RL training in the mathematical domain, their findings are limited. Our work offers a more thorough and systematic investigation of how PRMs can be leveraged in RL for code generation tasks.

\section{Conclusions and Limitations}

In this work, we tackled the challenge of sparse reward signals in RL for code generation by introducing a PRM that provides dense, line-level feedback. This approach, inspired by human-like code refinement, enhances learning efficiency and encourages better exploration. Our experiments show that integrating PRMs significantly improves the pass rates of code generation models across HumanEval, MBPP, and LiveCodeBench. Notably, PRM not only facilitates more effective RL training but also improves performance in long-horizon code generation scenarios.

Despite these promising results, our approach has several limitations that warrant further exploration. First, PRM effectiveness depends on the quality of the collected data. While we automate data collection using binary search and unit tests, this method may overlook nuances of code correctness and introduce noise, particularly in complex or ambiguous programming tasks. Second, despite using binary search to reduce overhead, PRM training remains computationally expensive. Third, our method relies on external verification (e.g., unit tests), which limits its applicability to domains lacking well-defined correctness criteria, such as creative writing or open-ended generation tasks.
Addressing these challenges presents exciting future research directions including improving PRM data collection strategies and exploring alternative evaluation methods to extend PRM applicability beyond structured domains like code generation.

\newpage

%\section*{Impact Statement}

%Recent advancements in large language models (LLMs) have significantly improved their ability to generate code, making them increasingly valuable for software development and related tasks. In this work, we propose an approach that leverages Partial Reward Models (PRMs) to refine LLM training, leading to more effective code generation. Our experiments demonstrate that PRM-based training improves model performance, particularly in long-horizon scenarios where intermediate feedback is essential. We believe that our method can enhance the reliability and utility of LLMs for real-world programming challenges, contributing to their broader adoption in software engineering and automation.

%While these improvements offer clear benefits, they also raise ethical and societal considerations. Enhanced code generation models may reduce barriers to software development, but they could also amplify issues such as bias in training data, over-reliance on AI-generated code, and potential vulnerabilities in automatically generated software. Developers and organizations should carefully evaluate AI-assisted coding solutions, ensuring proper oversight and validation of generated outputs.


%\sloppy

\bibliography{references}
\bibliographystyle{icml2025}
%\fussy

%%%%%%%%%%%%%%%%%%%%%%%%%%%%%%%%%%%%%%%%%%%%%%%%%%%%%%%%%%%%%%%%%%%%%%%%%%%%%%%
%%%%%%%%%%%%%%%%%%%%%%%%%%%%%%%%%%%%%%%%%%%%%%%%%%%%%%%%%%%%%%%%%%%%%%%%%%%%%%%
% APPENDIX
%%%%%%%%%%%%%%%%%%%%%%%%%%%%%%%%%%%%%%%%%%%%%%%%%%%%%%%%%%%%%%%%%%%%%%%%%%%%%%%
%%%%%%%%%%%%%%%%%%%%%%%%%%%%%%%%%%%%%%%%%%%%%%%%%%%%%%%%%%%%%%%%%%%%%%%%%%%%%%%
\newpage
\appendix
\onecolumn
%\newpage

%\appendix

\section{A Typical Example of the Learned Line-wise Rewards}

\begin{figure*}[ht]
\centering
\hspace{-8pt}

\resizebox{.95\textwidth}{!}{
\begin{tabular}{@{}c@{}}
    % First image spanning both columns
    \includegraphics[width=0.995\textwidth]{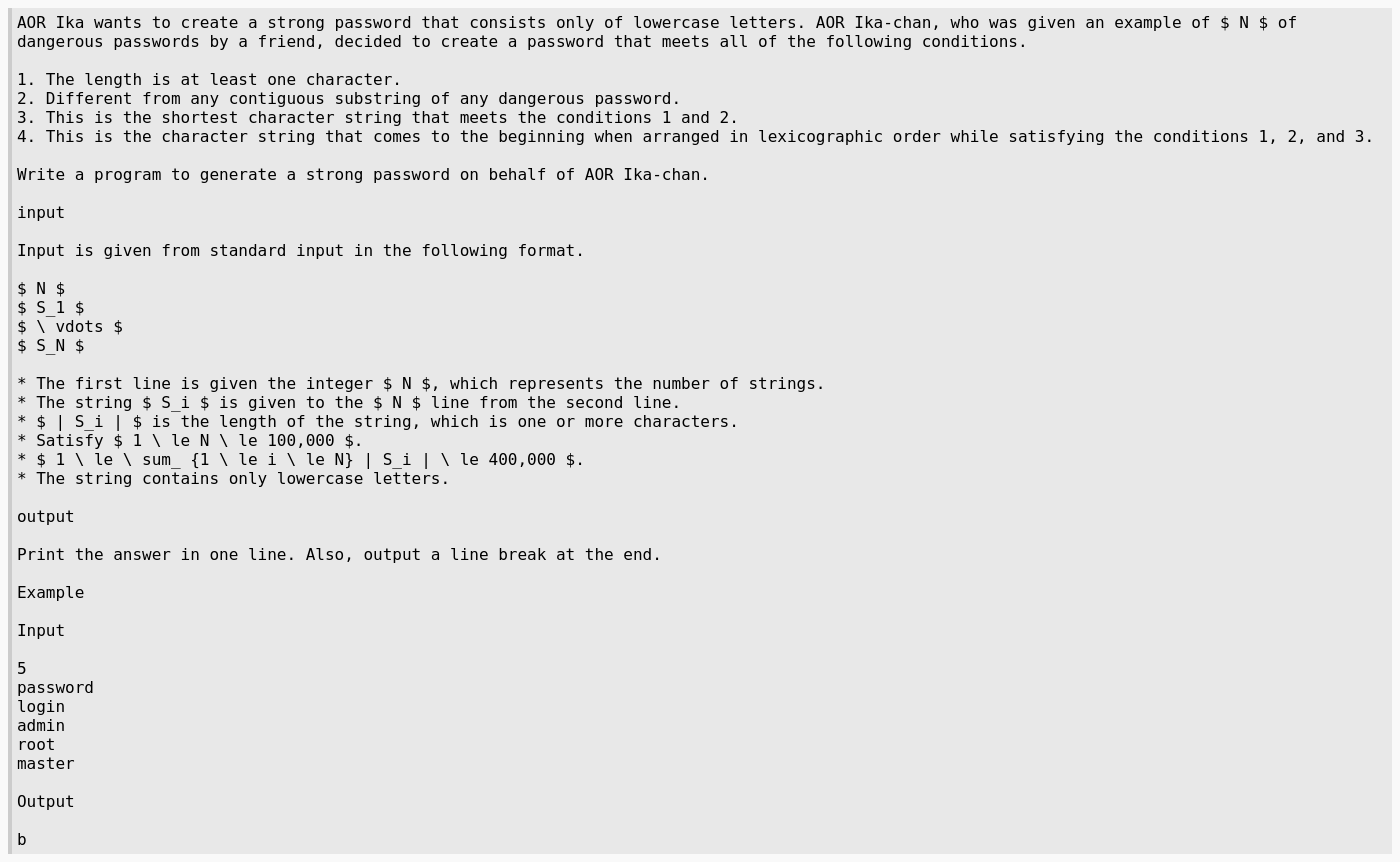} \\[-3pt]
    \begin{tabular}{l|l}
        % Titles for the second row images
        \hspace{3pt}\footnotesize{Line-wise Rewards Identified by Binary Search} & \hspace{-2pt}\footnotesize{Line-wise Rewards Predicted by a Learned PRM} \\
        % Second row: Two images side by side
        \includegraphics[width=0.4925\textwidth]{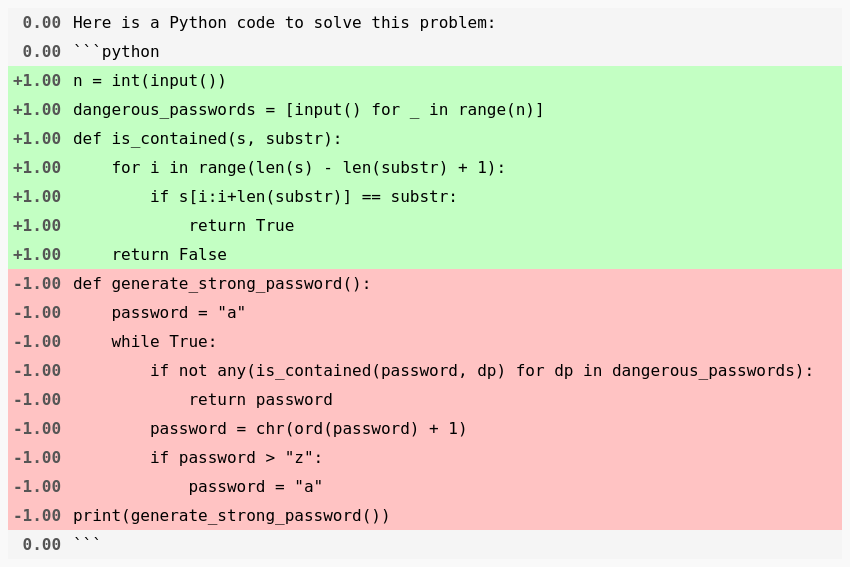}\hspace{-6pt} & \hspace{-6pt}\includegraphics[width=0.4925\textwidth]{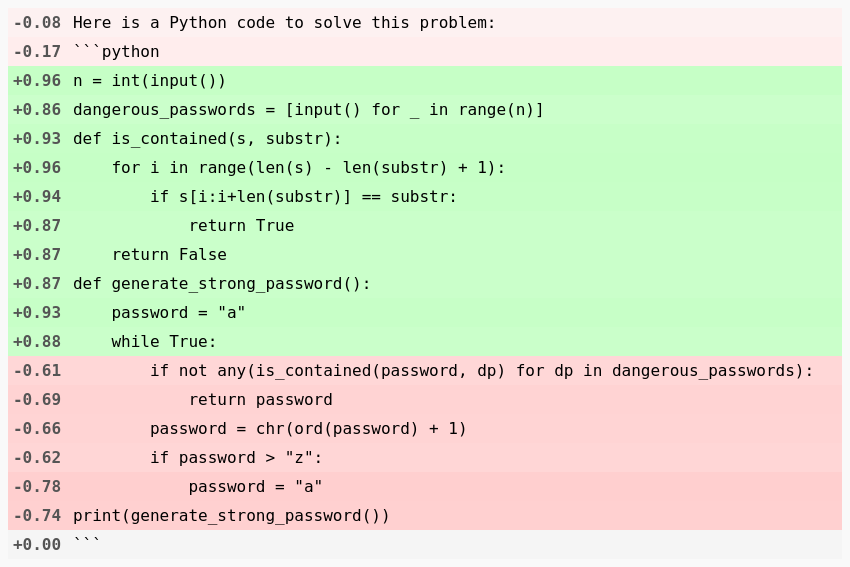} \\
    \end{tabular} \\
\end{tabular}
}
\caption{Visualization of the learned line-wise rewards. The top gray block displays the problem description, while the bottom section shows a model-generated response with line-wise rewards from different sources. The bottom-left block presents the line-wise rewards identified by binary search, and the bottom-right block presents the line-wise rewards predicted by a learned PRM. The actual reward value is shown at the beginning of each line, and each line is color-coded based on the reward value: lines with rewards closer to -1 are shaded \textcolor{red!60}{red}, while those closer to +1 are shaded \textcolor{green!70}{green}.}
\label{fig:linewise-rewards}
\end{figure*}

In Figure~\ref{fig:linewise-rewards}, we present a typical example of the line-wise rewards identified by binary search and predicted by a learned PRM to give readers a clearer understanding of our method. In this example, we first sampled a problem from the training set and used our in-house model to generate a response for it. For this generated response (which is not included in the PRM training data), we show the line-wise rewards derived from two sources:
\vspace{-3pt}
\begin{enumerate}[left=5pt]
    \item \textbf{Line-wise Rewards Identified by Binary Search:} We directly applied the model to perform Algorithm~\ref{alg:binary_search}, labeling the reward for each line.
    \item \textbf{Line-wise Rewards Predicted by a Learned PRM:} We used the learned PRM to predict the rewards for each line.
\end{enumerate}

\newpage

\section{RL Training Curves}

\begin{figure*}[th]
    \centering
    \includegraphics[width=\textwidth, trim={0 0 0 20pt}, clip]{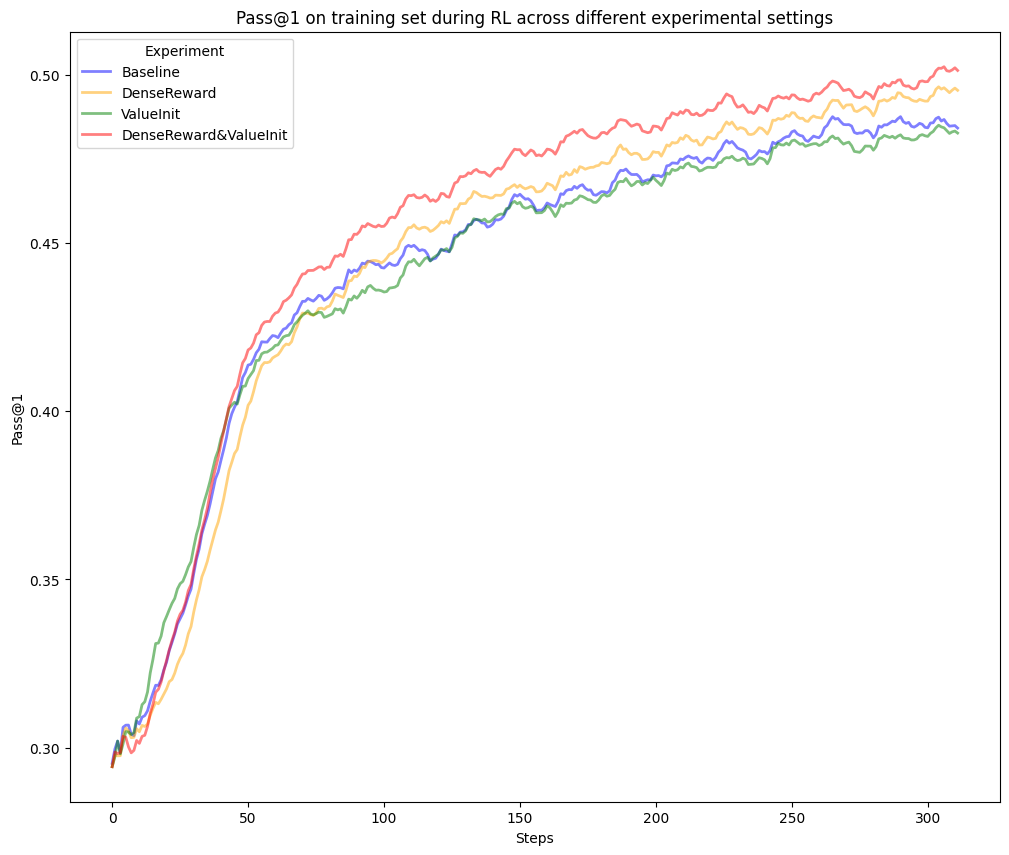}  
    \caption{RL training curve of all experiment settings in \texttt{Doubao-Lite-RL} experiments. Using PRM as both DenseReward and ValueInit (DenseReward\&ValueInit) yields the best result.}
    \label{fig:rl_curve}
\end{figure*}

In Figure~\ref{fig:rl_curve}, we present the smoothed RL training curves for all four settings (with and without DenseReward, and with and without ValueInit) in \texttt{Doubao-Lite-RL} experiments, using a moving average to reduce noise and enhance readability. These curves correspond to all four RL settings reported in Table~\ref{tab:performance}. The smoothed trends clearly show that when PRM is used as DenseReward, the model solves more problems compared to the baseline, demonstrating PRM's role in enabling more efficient exploration during RL training. Furthermore, when PRM is applied as both DenseReward and ValueInit, our method achieves the best performance.

\newpage

\section{PRM Training Data Statistics}

We present detailed statistics on the PRM datasets used in our experiments to evaluate the optimal PRM data selection strategy, as discussed in Section~\ref{sec:exp:main_results}. The experiments were conducted using the \texttt{Doubao-Lite} series models. Table~\ref{tab:prm_stat} summarizes the following key metrics: the number of prompt-response pairs (\textbf{\#Samples}); the total number of tokens across all responses (\textbf{\#Tokens}); the average number of lines in all responses (\textbf{Avg. \#Lines}); and the distribution of PRM labels (\verb|-1/0/+1|).

\begin{figure*}[th]
    \centering
    \includegraphics[width=.92\textwidth]{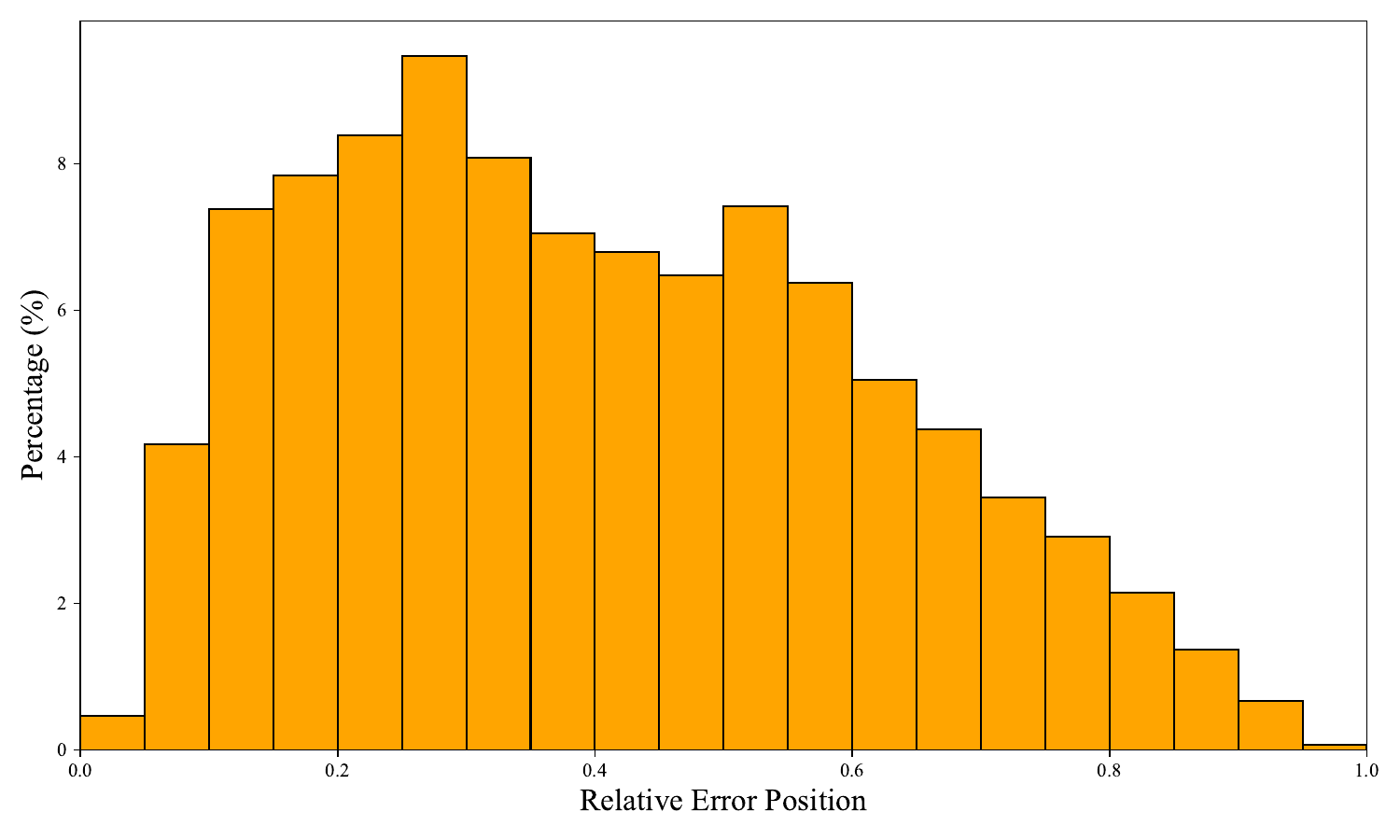}  
    \caption{Distribution of Relative Error Positions Identified by Binary Search.}
    \label{fig:error_dist}
\end{figure*}

In Figure~\ref{fig:error_dist}, we present the distribution of error positions of all {\textcolor{DarkOrange}{Revised}} responses (responses that initially fail but have a correct prefix identified) as determined by the Binary Search procedure (Algorithm~\ref{alg:binary_search}). The absolute error position (i.e., the position of the first token rejected by Binary Search) is normalized as follows: for a response $\mathbf{y} = (y_{1}, y_{2}, \ldots, y_{L})$ with $L$ tokens, if the Binary Search accepted the prefix $(y_{1}, y_{2}, \ldots, y_{p})$ consisting of $p$ tokens, the Relative Error Position is calculated as $\frac{p}{L}$.

\begin{table*}[h]
    \centering
    \caption{Statistics of PRM training data collected using different data selection strategies.}
    \vspace{0.1in}
    \begin{tabular}{lcccccc}
    \toprule
    \multirow{2}{*}{\textbf{Strategy}} & \multirow{2}{*}{\textbf{\#Samples}} & \multirow{2}{*}{\textbf{\#Tokens}} & \multirow{2}{*}{\textbf{Avg. \#Lines}} & \multicolumn{3}{c}{\textbf{PRM Labels}} \\ 
    \cmidrule(lr){5-7}
     & &  &  & \texttt{-1} & \texttt{0} & \texttt{+1} \\ 
    \midrule
    \textbf{Full}                   & 838K  & 179M   & 16.82   & 44.25\%   & 17.87\%   & 37.88\%   \\ 
    \textbf{Remove \textcolor{BrickRed}{Hard}} & 630K & 119M   & 15.06   & 24.03\%   & 19.18\%   & 56.79\%   \\ 
    \textbf{\textcolor{Goldenrod}{Medium} Only} & 485K & 104M   & 16.57   & 27.66\%   & 19.41\%   & 52.93\%   \\ 
    \textcolor{DarkOrange}{Revised} \textbf{Only} & 352K & 76M   & 16.71   & 13.20\%   & 19.42\%   & 67.38\%   \\ 
    \bottomrule
\end{tabular}
    \label{tab:prm_stat}
\end{table*}

\newpage

\section{Qwen2.5-7B Experiment Details}

\paragraph{Base Model.} We adopt \textbf{Qwen2.5-7B} as our base model~\citep{qwen2.5}, a recently released causal language model available at 
\href{https://huggingface.co/Qwen/Qwen2.5-7B}{\url{https://huggingface.co/Qwen/Qwen2.5-7B}}. Qwen2.5 belongs to the Qwen series of large language models~\citep{qwen2}, known for their advanced capabilities across a wide range of domains. The Qwen2.5-7B model has 7.61 billion total parameters (6.53 billion excluding embeddings) and utilizes the Transformer architecture as its core. It incorporates state-of-the-art enhancements, including Rotary Positional Embedding (RoPE), SwiGLU activation, RMSNorm, and Attention QKV bias. The model consists of 28 layers and employs 28 attention heads for queries (Q) and 4 for keys and values (KV), making it highly efficient for tasks requiring robust attention mechanisms.

\paragraph{SFT Settings.} We fine-tuned the Qwen2.5-7B model on the SFT dataset described in Section~\ref{sec:exp:setup}. The model was trained for two epochs, starting with a learning rate of \(1 \times 10^{-7}\), which linearly increased to \(2 \times 10^{-5}\) during the first 2\% of the total training steps. After reaching the peak learning rate, a cosine learning rate decay schedule was applied, gradually reducing the learning rate to \(2 \times 10^{-6}\) for the remainder of the training. Additionally, a constant weight decay of 0.01 was used throughout the SFT training process to regularize the model and improve generalization. The model fine-tuned through this process is referred to as \textbf{Qwen2.5-7B-SFT}.

\paragraph{RL Baseline.} We adopted the same RL baseline training method and used the same RLHF dataset described in Section~\ref{sec:exp:setup} to further train the Qwen2.5-7B-SFT model. For PPO training, we configured the following hyperparameters: a batch size of 4096, a linear warmup over the first 5 steps, followed by a constant learning rate of \(2 \times 10^{-6}\) for both the actor and critic, and a KL penalty of 0.01. The training utilized the AdamW optimizer and spanned approximately 300 steps, during which we empirically observed performance convergence.

\paragraph{PRM Training.} We selected four checkpoints at 50, 100, 150, and 200 steps during the training process of the RL baseline model. For each checkpoint, we sampled $n=5$ responses for every coding prompt in the training dataset \(\mathcal{D}_{\text{train}}\). Each sampled response was labeled using the binary search procedure described in Algorithm~\ref{alg:binary_search}, with \(K = 20\) completions generated for each partial code prefix.
The data collected from all checkpoints was then aggregated to form a PRM training set, employing the \textbf{\textcolor{DarkOrange}{Revised} Only} strategy described in Section~\ref{sec:exp:main_results}. This resulted in 165K samples and 28M tokens. On average, each response contained 16.07 lines. The PRM label distribution was 25.88\% for \verb|-1|, 15.90\% for \verb|0|, and 58.22\% for \verb|+1|. The PRM was initialized using the value model from the RL baseline and fine-tuned on this PRM dataset using the objective function defined in~\eqref{eq:prm_obj}.

\paragraph{Integrating PRM into RL.} We used the same settings and hyperparameters as described in Section~\ref{sec:exp:setup}. Additionally, we observed that due to the properties of the Qwen2.5-7B tokenizer, a newline token is not always represented as a simple \verb|"\n"| token. Instead, the tokenizer combines other non-space characters with an ending \verb|"\n"| to form new tokens (e.g., \verb|":\n"|, \; \verb|"):\n"|, \; \verb|")\n"|, \; \verb|"\n\n"|, \; \verb|"())\n"|, \verb|"]\n"|, \; \verb|"()\n"|, \; \verb|"():\n"|, etc.). This makes it more challenging to accurately identify line separator tokens in the model's responses.

Empirically, we addressed this challenge by selecting the 50 most frequent tokens in the PRM dataset whose corresponding token strings include \verb|"\n"|. The full list of token ids is shown below:
\[
\text{\{198, 510, 982, 340, 271, 2398, 921, 741, 3932, 1171, 692, 1305, 4167, 2546, 1447, 10343, 1138,  19324, }
\]
\[
\text{341, 5563, 9957, 382, 3407, 3646, 624, 48443, 280, 456, 2533, 3989, 1248, 5613, 8389, 8997, 698, }
\]
\[
\text{24135, 317, 7368, 2440, 10907, 22165, 4432, 5929, 7129, 345, 11043, 532, 4660, 21686, 14288\}.}
\]

During RL training, we only applied partial rewards from PRM to these tokens.

\end{document}